\documentclass{sig-alternate-2013}
\usepackage{color}
\usepackage{hyperref}
\usepackage{url}
\usepackage{verbatim} 
\usepackage{subfigure} 
\usepackage{multirow}
\usepackage{algorithm}
\usepackage[noend]{algorithmic}
\usepackage{xspace}
\usepackage{graphicx}
\usepackage{epsfig}
\usepackage{amsmath}
\usepackage{amssymb}
\usepackage{times}
\usepackage{booktabs} 
\usepackage{enumitem}

\newcommand*\samethanks[1][\value{footnote}]{\footnotemark[#1]}

\newcommand{\cut}[1]{}
\newcommand{\xhdr}[1]{\vspace{1.7mm}\noindent{{\bf #1.}}}

\newcommand{\eg}{{e.g.}}

\graphicspath{{./FIG/}}

\permission{Copyright is held by the International World Wide Web Conference Committee (IW3C2). IW3C2 reserves the right to provide a hyperlink to the author's site if the Material is used in electronic media.}
\conferenceinfo{WWW 2015,}{May 18--22, 2015, Florence, Italy.}
\copyrightetc{ACM \the\acmcopyr}
\crdata{978-1-4503-3469-3/15/05. \\
http://dx.doi.org/10.1145/2736277.2741688}

\begin{document}

\title{QUOTUS: The Structure of Political Media Coverage\\
as Revealed by Quoting Patterns}

\numberofauthors{5}
\author{
\alignauthor 
    Vlad Niculae\thanks{The first three authors contributed equally and are ordered alphabetically. The last two authors are also ordered alphabetically.}  \thanks{The research described herein was conducted in part while these authors were at the Max Planck Institute for Software Systems.} \\
    \affaddr{
      Cornell University
      \email{vlad@cs.cornell.edu}
    } 
    \alignauthor 
    Caroline Suen\samethanks[1]\\
    \affaddr{
      Stanford University
      \email{cysuen@stanford.edu}
    } 
        \alignauthor 
    Justine Zhang\samethanks[1] \samethanks[2]\\
    \affaddr{
      Stanford University
      \email{justinez@stanford.edu}
    } 
            \and 
        \alignauthor \hspace*{-1cm} 
    \mbox{Cristian Danescu-Niculescu-Mizil\samethanks[2]}\\
    \affaddr{
      Cornell University
      \email{cristian@cs.cornell.edu}
    }
    \alignauthor
       Jure Leskovec\\
    \affaddr{
      Stanford University
      \email{jure@cs.stanford.edu}
    } 
}

\maketitle


\newcommand{\liberal}{liberal\xspace}
\newcommand{\conservative}{conservative\xspace}

\newcommand{\independent}{independent\xspace}
\newcommand{\mainstream}{mainstream\xspace}

\newcommand{\liberals}{liberals\xspace}
\newcommand{\conservatives}{conservatives\xspace}

\newcommand{\Liberal}{Liberal\xspace}
\newcommand{\Conservative}{Conservative\xspace}

\newcommand{\sC}{sC\xspace}
\newcommand{\suspectedconservative}{suspected conservative\xspace}
\newcommand{\SuspectedConservative}{Suspected Conservative\xspace}
\newcommand{\dC}{dC\xspace}
\newcommand{\declaredconservative}{declared conservative\xspace}
\newcommand{\DeclaredConservative}{Declared Conservative\xspace}

\newcommand{\sL}{sL\xspace}
\newcommand{\suspectedliberal}{suspected liberal\xspace}
\newcommand{\SuspectedLiberal}{Suspected Liberal\xspace}
\newcommand{\dL}{dL\xspace}
\newcommand{\DeclaredLiberal}{Declared Liberal\xspace}
\newcommand{\declaredliberal}{declared liberal\xspace}

\newcommand{\propensity}{proportion\xspace}
\newcommand{\Prop}{M\xspace}

\newcommand{\sectionrule}{\addlinespace[1.5ex]}
\newcommand{\smallrule}{\addlinespace[0.5ex]}

\newcommand{\quotusvizurl}{http://snap.stanford.edu/quotus/vis}

\newcommand{\catA}{A\xspace}
\newcommand{\catB}{B\xspace}

\begin{abstract}


Given the extremely large pool of events and stories available, media outlets
need to focus on a subset of issues and aspects to convey to their audience.
Outlets are often accused
of exhibiting
a systematic bias in this selection
process, with different outlets portraying different versions of
reality.
However, in the absence of objective measures and empirical evidence, the
direction and extent of systematicity remains widely disputed.

In this paper we propose a framework based on quoting patterns for quantifying and characterizing the degree to which media outlets exhibit systematic bias.  We apply this framework to a massive dataset of news articles spanning the six years of Obama's presidency and all of his speeches, and reveal that a systematic pattern does indeed emerge from the outlet's quoting behavior.
Moreover, we  show that this pattern can be successfully exploited in an unsupervised prediction setting, to determine which new quotes an outlet will select to broadcast.
By encoding bias patterns in a low-rank space we provide an analysis of the structure of political
media coverage.
This reveals a latent media bias space that aligns surprisingly well with political ideology and outlet type.
A linguistic analysis 
exposes striking differences across these latent dimensions, showing how the different types of media outlets portray different realities even when reporting on the same events.  
For example, outlets mapped to the \mainstream \conservative side of the latent space focus on
quotes that portray a presidential persona disproportionately characterized by negativity.

\end{abstract}

\vspace{1mm}
\noindent {\bf Categories and Subject Descriptors:} H.2.8 {\bf
[Database Management]}: Database applications---{\it Data mining}

\noindent {\bf General Terms:} Algorithms; Experimentation.


\vspace{0.2in}
\section{Introduction}
\label{sec:intro}


The public relies heavily on mass media outlets for accessing important information on current events. 
Given the intrinsic space and time constraints these outlets face,
some
filtering of events, stories and aspects to broadcast is unavoidable.  

The majority of media outlets claim to be balanced in their coverage, selecting issues purely based on their newsworthiness.
However, journalism watchdogs  
 and political think-tanks
 often accuse outlets of exhibiting 
  systematic bias in the selection process, leaning either towards serving the interests of the owners and journalists or towards appeasing the preferences of their intended audience.  This phenomenon, generally called media bias, has been extensively studied in political science, economics and communication literature (for a survey see \cite{Prat::2013}).   Theoretical accounts provide a taxonomy of media bias based on the level at which the selection takes place: e.g., what issues and aspects are covered (issue and facts bias), how facts are presented (framing bias) or how they are commented on %
(ideological stand bias).  Importantly, the dimensions along which bias operates can be very diverse: although the most commonly discussed dimension aligns with 
political ideology (e.g., liberal vs. conservative bias) other dimensions such as mainstream bias, corporate bias, power bias, and advertising bias are also important and perceived as inadequate journalism practices.

 \begin{figure*}[t]
     \centering
     \includegraphics[width=1\textwidth]{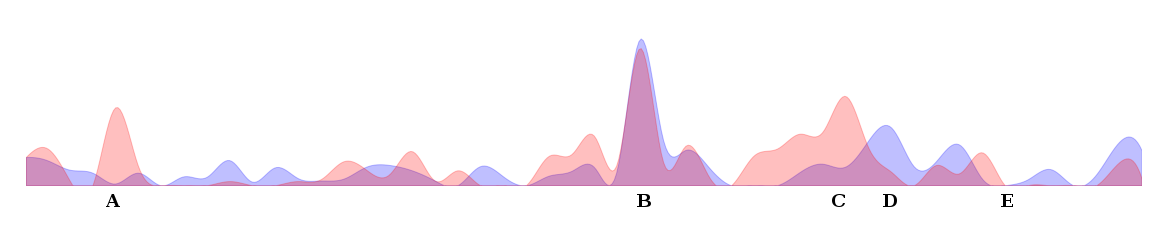}
     \caption{Volume of quotations for each word from a fragment of the 2010 State of the Union Address split by political leaning: \conservative outlets shown in red and \liberal outlets shown in blue. Quotes from the marked positions are reproduced in Table \ref{tab:example} and shown in
the
QUOTUS visualization in Figure \ref{fig:topic}.
 }
     \label{fig:example}
 \end{figure*}

\begin{table*}[t]
    \centering
\begin{tabular}{c p{15cm}}
    \toprule
        \textbf{Position} & \textbf{Quote from the 2010 State of the Union Address} \\
        \midrule
A & {And in the last year, hundreds of al Qaeda's fighters and affiliates, including many senior leaders, have been captured or killed---far more than in 2008.} \\
\sectionrule 
B & {I will work with Congress and our military to finally repeal the law that denies gay Americans the right to serve the country they love because of who they are. It's the right thing to do.}                                                \\
\sectionrule 
C & {Each time lobbyists game the system or politicians tear each other down instead of lifting this country up, we lose faith. The more that TV pundits reduce serious debates to silly arguments, big issues into sound bites, our citizens turn away.} \\
\sectionrule
D & {Democracy in a nation of 300 million people can be noisy and messy and complicated. And when you try to do big things and make big changes, it stirs passions and controversy. That's just how it is. }                                                                                   \\
\sectionrule
E & {But I wake up every day knowing that they are nothing compared to the setbacks that families all across this country have faced this year. }                                                                                             \\
        \bottomrule
    \end{tabular}
\caption{Quotes corresponding to the positions marked in Figure \ref{fig:example}.}
\label{tab:example}
\end{table*}

Bias is a highly subjective phenomenon that is hard to quantify---something that is considered unfairly biased
by some might be
regarded as balanced by others.  For example, a recent Gallup survey~\cite{Gallup:2010} shows not only that the majority of Americans (57\%) perceive media as being biased, but also that the perception of bias varies vastly depending on their self-declared ideology: 73\% of conservatives perceive the media as having a liberal bias, while only 11\% of liberals perceive it as having a liberal bias (and 33\% perceive it as having a conservative bias). %
As a consequence, the extent and direction of bias for individual media outlets remains highly disputed.\footnote{In response to numerous accusations, the 21st Century Fox CEO Rupert Murdoch has declared ``I challenge anybody to show me an example of bias in Fox News Channel.'' (Salon, 3/1/01).}

The subjective nature of this phenomenon and the absence of large scale objectively labeled data hindered
quantitative analyses~\cite{Ho:QuarterlyJournalOfPoliticalScience:2008}, and consequently most 
existing empirical studies of media bias are small focused analyses
\cite{dalton1998partisan,peake2007presidents,Schiffer:PoliticalCommunication:2006}.
 A few notable computational studies circumvent these limitations by relying on proxies such as the similarity between media outlets and the members of congress
 \cite{Gentzkow:Econometrica:2010,Groseclose01112005,lin2011more}
 or U.~S. Supreme Court Justices \cite{Ho:QuarterlyJournalOfPoliticalScience:2008}.
Still, the reliance on such proxies constrains the analysis to predetermined dimension of bias, and
conditions the value of the results on the accuracy of the proxy assumptions
\cite{gasper2011shifting,Liberman:2005}.

\xhdr{Present work: unsupervised framework}
We
present a framework for quantifying the systematic bias exhibited by media outlets, without relying on
any annotation %
and without predetermining the dimensions of bias.  The basic operating principle of this framework is that quoting patterns exhibited by individual outlets can reveal media bias.
Quotes are especially suitable since they correspond to an outlet's explicit choices of whether to cover or not
specific
parts a larger statement.
In this sense, quotes have the potential to provide precise insight into the decision process behind media coverage.

Ultimately, the goal of the proposed framework is to quantify to what extent quoting decisions follow systematic patterns that go beyond the relative importance (or newsworthiness) of the quotes, and to characterize the dimensions of this bias. %

As a motivating example, Figure \ref{fig:example} illustrates media quotations from a fragment of the U.S. President, Barack Obama's 2010 State of the Union Address. 
The text is ordered along the
{\em x} axis, and the {\em y} axis
corresponds to the number of times a particular part of the Address was quoted.
We display quoting volume by outlets considered\footnote{While our core methodology is completely unsupervised and not limited to the \conservative--\liberal direction, we use a small set of manual labels for interpretation (as detailed in Section \ref{sec:experiments}).} 
to be \liberal (blue) or \conservative (red).  We observe both similarities and differences in what parts of the
address
get quoted. For example, the quote at position B (reproduced in Table \ref{tab:example}) is highly cited by both sides, while the quote at position A is cited by \conservative outlets and largely ignored by \liberal outlets.\footnote{We invite the reader to explore more such examples using the online visualization tool we release with this paper:
\url{\quotusvizurl}}
To a certain extent, the audience of the \liberal media experienced a different State of the Union Address than the audience of the \conservative group.  %
Are these variations just random fluctuations,
or are they the result of a systematically biased selection process?  Do different media outlets portray consistently different realities even when reporting on the same events? %

To study and identify the presence of systematic bias at a large scale, we
start  from a massive collection of six billion news articles~\cite{memetracker2009} (over 20TB of
compressed data) which spans the
six years
of Barack Obama's tenure in office as the President of the United States (POTUS), between 2009 and 2014.  We use the 2,274 public speeches made by Obama during this period (including state of the union addresses, weekly presidential addresses and various press
conferences). 
We match quotes from Obama's speeches to our news articles and build an outlet-to-quote bipartite graph linking
275
media outlets to the 
over 267,000 quotes which they reproduce.  The graph allows us to study the structure of political media coverage over a long period of time and over a diverse set of public issues,
while at the same time maintaining uniformity with respect to the person who is quoted.

\xhdr{Focused analysis} Before applying our unsupervised framework to the entire data, we first perform a 
small-scale
focused analysis on a carefully constructed subset of labeled outlet
leanings, in order to gain intuition about the nature of the data. 
We label outlets based on \liberal and \conservative leaning, as well as on whether or not their leaning is self-declared. 
An empirical investigation of various 
characteristics of the outlets and of their articles
 reveals differences 
 that  further
 motivate
 the need for an unsupervised approach that is not tied to a predetermined dimension of bias.

\xhdr{Large-scale analysis}  To quantify the degree to which the outlet-to-quote bipartite graph encodes a systematic pattern that extends beyond the simple newsworthiness of a quote,
we use an unsupervised prediction paradigm.\footnote{Our approach is unsupervised in the sense of not using any annotation
or prior knowlegde about the bias or leaning of either news outlets or quote
content.
}
    The task is to predict whether a given outlet will select a new
    quote, based
    on its previous quoting pattern.  We show that, indeed, the patterns encoded in the outlet-to-quote graph can be exploited efficiently via a matrix factorization approach akin to that used by recommender systems.  Furthermore, this approach brings significant improvement over baselines that only encode the popularity of the quote and the propensity of the outlet to pick up quotes, showing that these can not fully explain the systematic pattern that drives content selection. 

Factorizing the outlet-to-quote matrix provides new insights into the structure of the political media coverage.  First, we find that our labeled \liberal and \conservative outlets are separated in the
space defined by the first two
latent dimensions.
Moreover, a post-hoc analysis of the outlets mapped to the extremes of this space reveals a strong alignment between these two latent dimensions and media type (\mainstream
vs.\ \independent)
and political ideology.  The separation between outlets along these dimensions is surprisingly clear, considering that the method is completely unsupervised.

By mapping the quotes onto the same latent space, our method also reveals how the systematic patterns of the media operate at a linguistic level.  For example, outlets on the \conservative side of the (latent) ideological spectrum are more likely to select Obama's quotes that contain more negations and negative sentiment, portraying an overly negative character.

To summarize the main contributions of this paper:
\begin{itemize}
\item we introduce a completely unsupervised framework for analyzing the patterns behind the media's selection of what to cover;
\item we apply this framework to a large dataset of presidential speeches and media coverage thereof which we make publicly available together with an online visualization tool (Section \ref{sec:proposed});
\item we reveal systematic biases that are predictive of an outlet's quoting choices (Section \ref{sec:matrixcompletion});
\item we show that the most important dimensions of bias align with the ideological spectrum and outlet type (Section \ref{sec:lowrank});
\item we characterize these two dimensions linguistically, exposing striking differences in the way in which different outlets portray reality (Section \ref{sec:analysis}).

\end{itemize}

\section{Methodology}
\label{sec:proposed}


 \begin{figure*}[ht]
     \centering
     \includegraphics[width=1\textwidth]{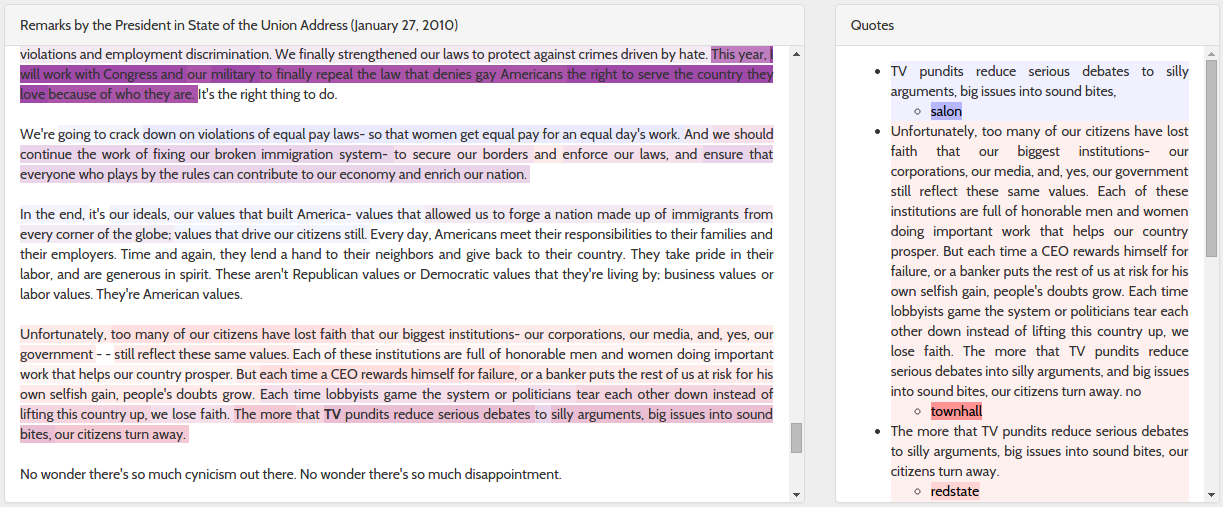}
     \caption{Visualization of the fragment of 2010 State of the Union Address represented in Figure \ref{fig:example} between markers B and C.  The left panel shows text highlighted according to quotation volume and slant.  The right panel shows all variants of a selected quote cluster.  An interactive visualization for the entire  dataset is available online  at {\small \url{http://snap.stanford.edu/quotus/vis/}.} 
 }
     \label{fig:topic}
 \end{figure*}

Our
analysis framework
relies on a bipartite graph that encodes various outlets' selections
of quotes to cite.
First, we introduce the general methodology for building 
such a bipartite graph from transcript and news article data. We then apply this methodology
to the particular setting in which we instantiate this graph: with
speeches delivered by President Obama
and a massive collection of news articles.

\subsection{Building an outlet-to-quote graph}\label{sec:graph}

\xhdr{Matching}
We begin with the two datasets that we wish to match: a set of source statements (in our case, presidential speech transcripts), and a set of
news articles. 
We identify the quotes in each article, and search for a
candidate speech and corresponding location 
within the speech from which the quote
originates.
We allow approximate matches and align article quotes to the speeches word by word.

In order to avoid false positives, we set a lower bound $l$ on the number of words required
in each quote. For each remaining quote, we then examine the speeches
that occur before the 
quote's corresponding article's timestamp to
find a match.
Since more recent speeches are more likely to be quoted, 
for performance reasons
we search the
latest
speech first,
and proceed backwards.
Because matches that are too distant in time are more likely to be spurious,
we also limit
the set of candidates
to speeches 
that occur at most timespan $t$ before the
quote.

We find approximate matches using a variant of the Needleman-Wunsch dynamic programming
algorithm for matching strings
using
substring edit distance. 
We use an empirically determined similarity
threshold $s$ to determine whether a quote matches or not.\footnote{In our implementation, we use $l=6$, $t=7$ days, and $s=-0.4$.} The output of our matching process is 
an alignment between article quotes and source statements from the transcripts.

\xhdr{Identifying quote clusters}
News outlets often quote the same part of a speech in
different ways.
Variations result when articles select different subparts of the same
statement,
or
choose different paraphrases of the quote.
Sometimes these variations can be semantic 
while most of the times they are purely syntactic (\eg, resolution of pronouns).
For our purposes, we want to consider all variations of a quote as a unique
quote phrase that the media outlets choose to distribute.
To accomplish this, we group two different quotes into the same
\textit{quote cluster} if their matched locations within a transcript overlap on at least
five words. 

The resulting output is a series of quote clusters, each of which is affiliated with
a specific area of the statement transcript.
The majority of our analysis considers quotes at the cluster level, instead of looking at 
individual quote variants.

\xhdr{Article deduplication}
To most clearly highlight any relationships
between quote selection and
editorial slant,
we wish to ensure that each quote used in analysis is
deliberately chosen by the
news outlet
that
published
the corresponding article.
However, it is a common practice among news outlets to republish content
generated by other organizations. Notably, most news outlets will frequently
post articles generated by wire services.
Such curated content, while endorsed by the outlet, is not necessarily reflective of the
outlet's editorial stance, and we can better differentiate outlet quoting
patterns after removing them.
Duplicate articles
need not
necessarily be perfectly identical, so we employ fuzzy
string matching using length-normalized Levenshtein edit distance.
Among each set of duplicate articles, we keep the one published first.

\xhdr{Outlet-to-quote bipartite graph}
The output after executing our pipeline above is a set of (outlet, quote
phrase) pairs.\footnote{Some news outlets cite the same quote multiple times
across different articles.  To minimize the effect of multiple quoting, we keep
only the chronologically earliest quote, and disregard the rest.}
As a final step, we turn 
these pairs into a directed bipartite graph $G$,
with outlets and quote clusters as the two disjoint node sets.
An edge $u \rightarrow v$ exists in $G$ if outlet $u$ has an article that
picks up a variant of quote $v$.
\vspace{0.5cm}
\subsection{Dataset description}
We construct a database of presidential speeches by crawling the archives of public broadcast transcripts from
the White House's web site.\footnote{\url{http://whitehouse.gov}}
In this way we obtain full transcripts of speeches delivered by White
House--affiliated personnel, spanning from 2009 to 2014.
For the purposes of our analyses, we focus on the paragraphs that are specifically spoken by President Obama.
Our news article collection consists of articles spanning 
from 2009 to 2014; each entry includes the article's title, timestamp,
URL, and content.
Overall the collection contains over six 
billion articles.
To work with a more manageable amount of data, we 
run our quote matching pipeline on articles only if they contain the string ``Obama'', and were produced by one of 275 
media outlets from a list that was manually compiled; news outlets
on this list were identified as likely to produce content related to politics.
Overall,
this reduces the collection to
roughly 200GB of compressed news article data.
Further statistics about the
processed
data
are displayed in 
Table~\ref{tab:datasize}.

\begin{table}
\centering
\begin{tabular}{l r}
	\toprule
	\text{Number of news outlets} & \text{275} \\
	\text{Number of presidential speeches} & \text{2,274} \\
	\text{Number of unique articles} & \text{222,240} \\
	\text{Number of unique quotes} & \text{267,737} \\
	\text{Number of quote clusters} & \text{53,504} \\
	\text{Number of unique (outlet, cluster) pairs} & \text{228,893}\\
	\bottomrule
\end{tabular}
\caption{Statistics of the news article and presidential speech dataset used.}
\label{tab:datasize}
\end{table}

\xhdr{Visualization} Finally, we make the matched data publicly available and provide
an online visualization to serve as an interface for qualitative investigations of the data.\footnote{\url{http://snap.stanford.edu/quotus/}}

Figure \ref{fig:topic}
shows
a screenshot of this visualization. 
Quoted
passages have been
color-coded
with
color
intensity corresponding to the
volume of quotation, and shade corresponding to
editorial
slant (described in Section \ref{sec:outlets}).
The right panel displays variations of a selected quote---grouped in one of the
quote clusters---as represented by various outlets.
Each quote is hyperlinked to the articles containing it.
We provide this visualization as a potential tool with which political scientists and other researchers can
develop more insight about the structure of political media coverage.

\section{Small-scale Focused Analysis}
\label{sec:experiments}


\begin{figure*}[ht]
\centering
\subfigure[]{
\centering
\includegraphics[width=0.235\textwidth]{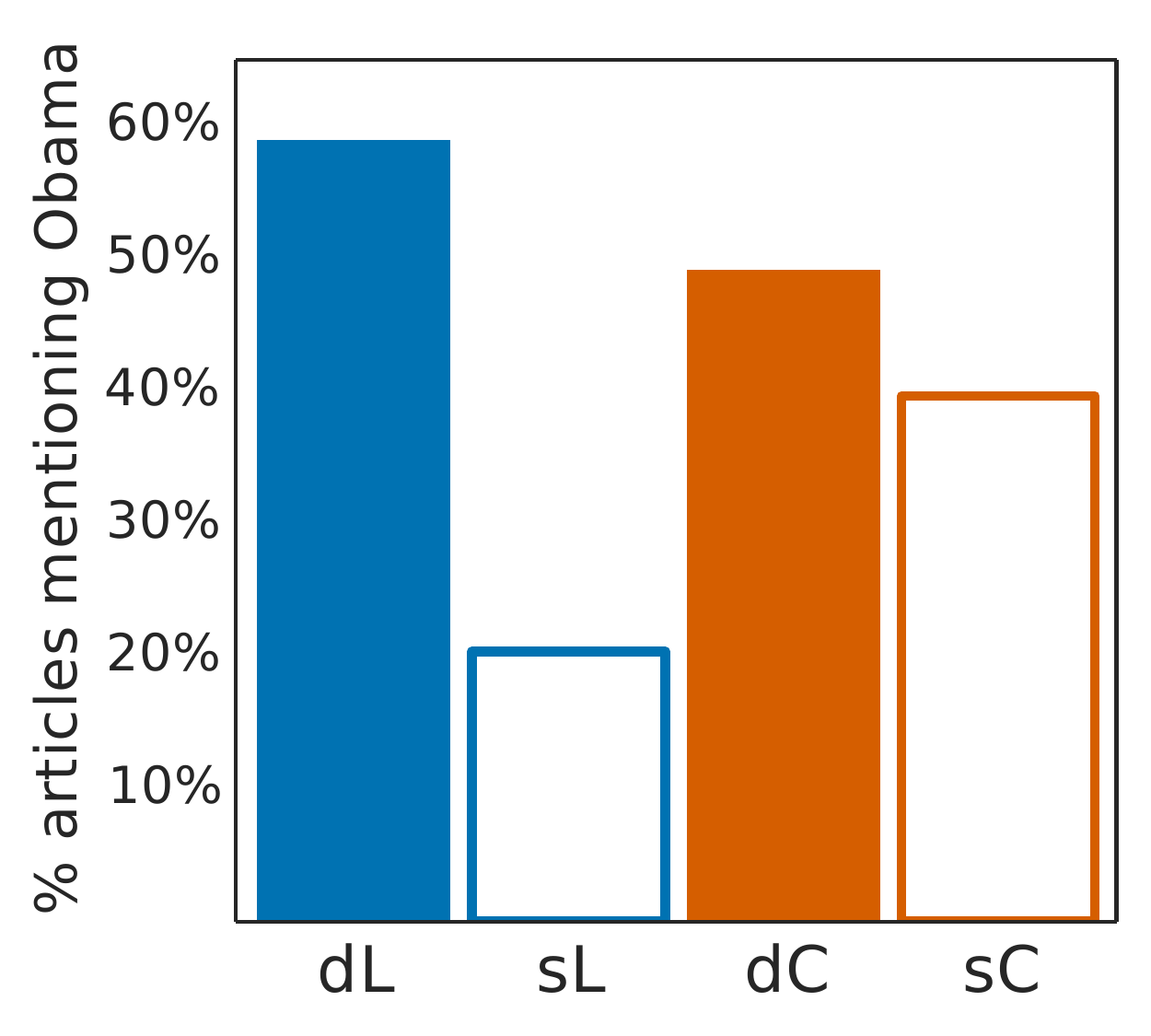}
\label{fig:mention_obama}
}
\subfigure[]{
\centering
\includegraphics[width=0.235\textwidth]{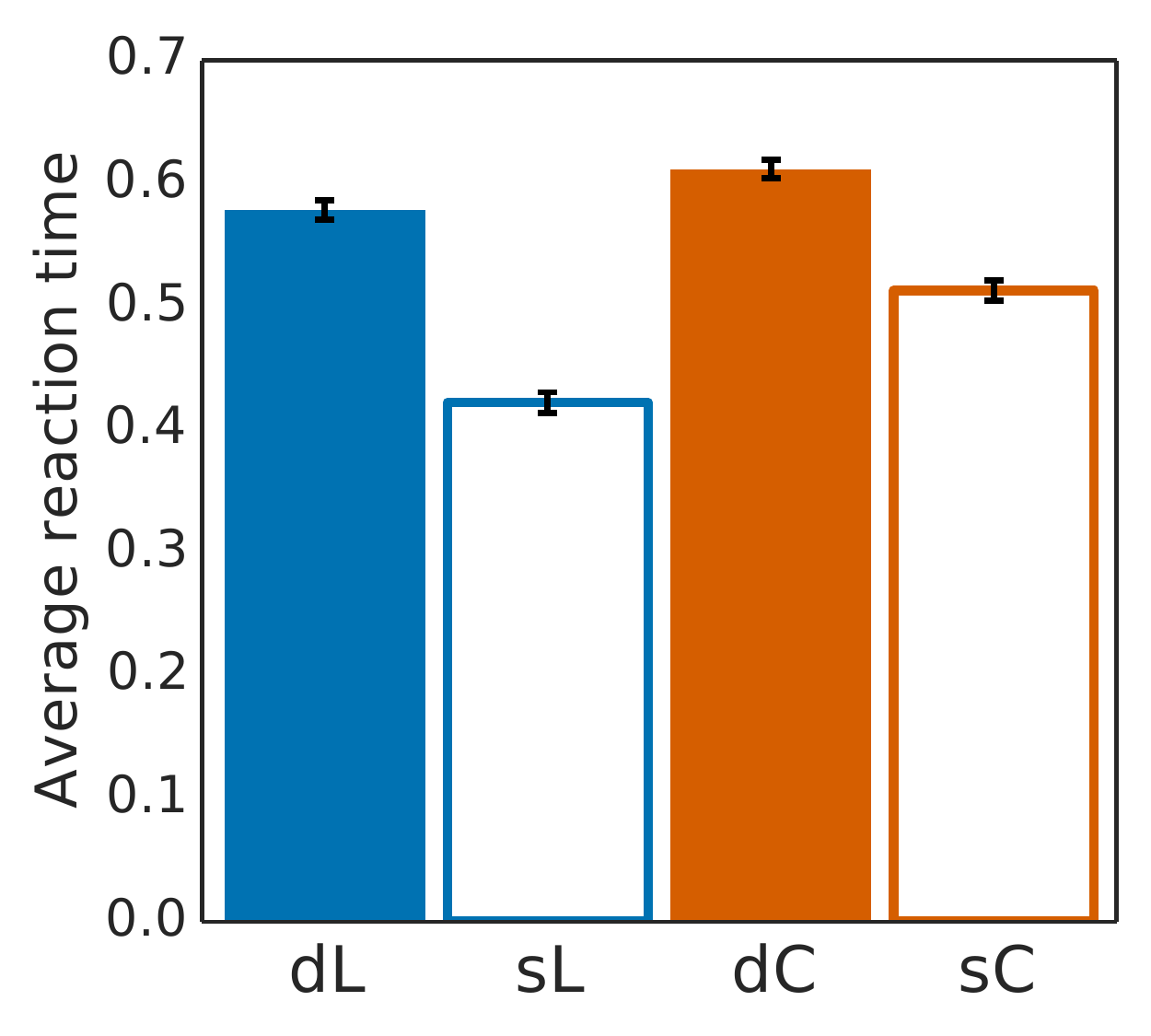}
\label{fig:reactiontime}
}
\subfigure[]{
\centering
\includegraphics[width=0.235\textwidth]{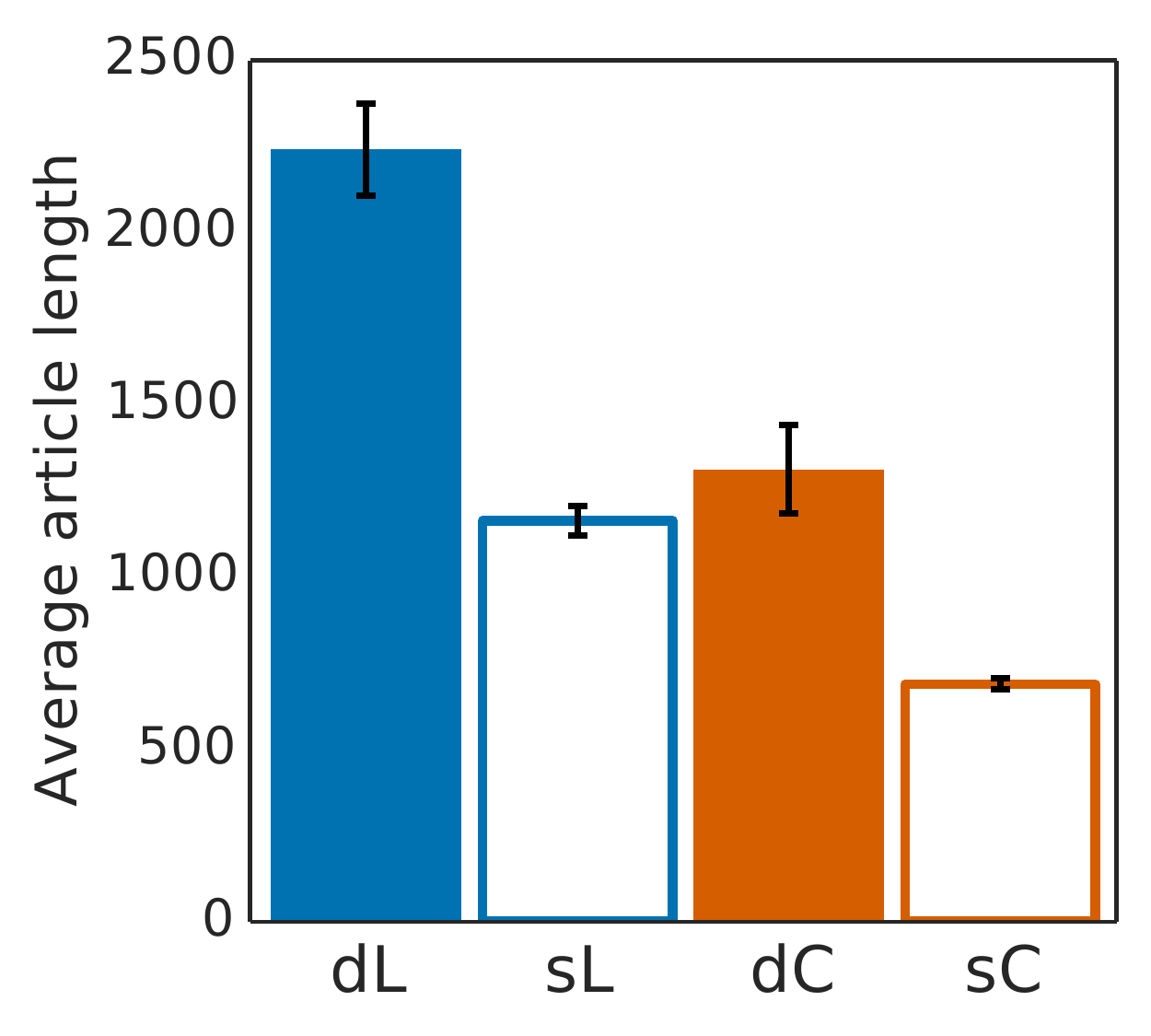}
\label{fig:avg_article_length}
}
\subfigure[]{
\centering
\includegraphics[width=0.235\textwidth]{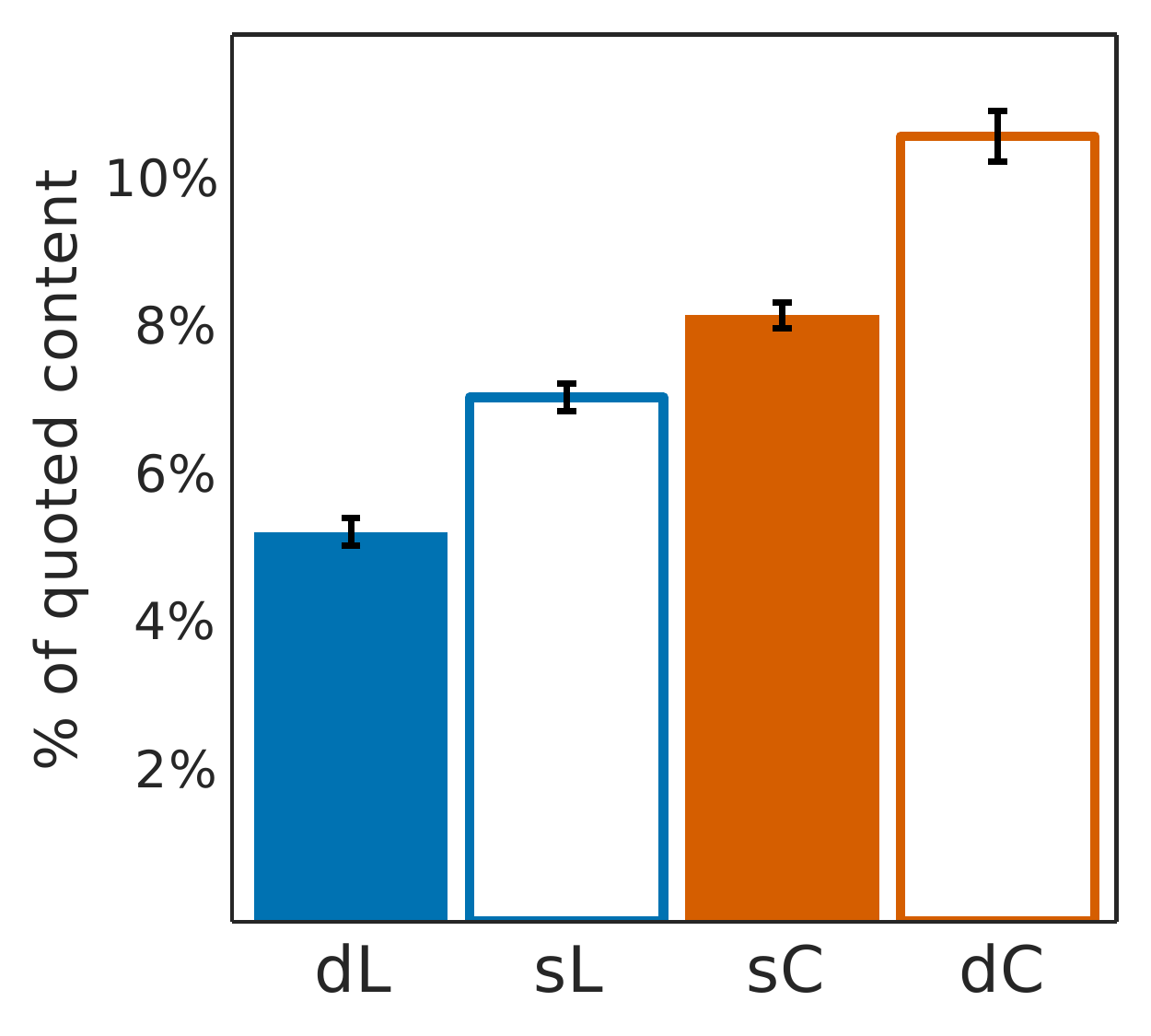}
\label{fig:quoted_content}
}
\caption{Differences between outlet categories: (a) the fraction of articles
mentioning the president; 
(b) the normalized reaction time to presidential speeches;
(c) the average word count of an article;
(d) the percentage of quoted content
in an article.
Filled bars correspond to outlets with declared political slants, and unfilled
bars to suspected ones.  Error bars indicate standard error.}
\label{fig:empirical}
\end{figure*}
To gain intuition about our data and to better understand the biases that occur in the political news landscape, we
first focus our analyses on a small subset of news outlets for which there is an established 
or suspected
bias
according to political science
research~\cite{baum2008new,comscore2010,Groseclose01112005}. 
We conduct an empirical analysis to 
compare outlets in different label categories in terms of their coverage of the presidential speeches.
Then, by  
interpreting this data
as a bipartite graph, we perform a rewiring experiment to
quantify the degree to which 
outlet categories relate to each other.  In doing so,  we motivate the need for an unsupervised approach to study the structure of political media bias at scale (Section \ref{sec:discussion}).

\subsection{Outlet selection}
\label{sec:outlets}

As discussed in the introduction, obtaining reliable labels of outlet political leaning is a challenge that has hindered quantitative analysis of this phenomenon.
One of the most 
common
 dichotomies considered in the literature is that  between \liberal and \conservative
ideologies.
We 
refer to political science research to construct a list of twenty-two outlets
which we group in four
  categories: \textit{\declaredconservative,}
\textit{\suspectedconservative,} \textit{\suspectedliberal,} and
\textit{\declaredliberal.} Our selection criteria is as follows:
\begin{itemize}[noitemsep]
  \item If an outlet declares itself to be liberal or conservative, or the owner explicitly declares a leaning, we refer to the outlet as \declaredconservative (\dC) or \declaredliberal (\dL). We refer to such outlets as \textit{declared outlets}.
  \item If several bias-related political science
studies~\cite{baum2008new,comscore2010,Groseclose01112005} consistently suggest that an outlet is \liberal or \conservative,      
but the outlet itself does not have a declared leaning, we label it as \suspectedliberal (\sL)  or \suspectedconservative (\sC). We refer to such outlets as \textit{suspected outlets}. 
\end{itemize}

\noindent The labeled outlets considered in this sections
 is shown in Table \ref{tab:outlets}.

\subsection{Outlet and article characteristics}
\label{sec:empiricalanalysis}
We first perform an empirical analysis to explore the relation between outlet categories.  We analyze both general characteristics of the outlets, as well as properties of the articles citing the president.

\xhdr{Percentage of articles discussing Obama}
First, we simply measure the percentage of all articles of a given outlet 
 that mention ``Obama''.
Figure~\ref{fig:mention_obama} shows
that the fraction of articles discussing the president is generally higher for 
 declared outlets (both \liberal and \conservative; filled bars) than for suspected ones (unfilled bars).
 This aligns with the intuition that outlets with a clear ideological affiliation are more likely to discuss political issues.

\xhdr{Reaction time}
Declared and suspected outlets also differ in how early they cover popular
speech segments. Many sound bites from Obama's more popular speeches,
such as the annual state of the union address, are cited by a multitude of
outlets. Here we consider  quotes that were cited by at least five of the
labeled outlets; for each such quote we sort the citing outlets into a relative
time ranking between 0 and 1, to normalize for cluster size, where a smaller
number indicates a shorter \textit{reaction time}. Aggregated results by outlet
category are shown in Figure
\ref{fig:reactiontime}.
We note that suspected outlets, especially liberal ones,
tend to report quotes faster
than those with declared ideology.

\begin{table}[t]
    \centering
    \begin{tabular}{l l}
    \toprule
        \textbf{\DeclaredConservative (\dC)} & \textbf{\DeclaredLiberal (\dL)} \\
    Daily Caller & Crooks and Liars  \\
    National Review & Daily Kos \\
    PJ Media  & Mother Jones \\
    Real Clear Politics & The Nation \\
    Reason & Washington Monthly \\
    The Blaze &  \\
    Weekly Standard & \textbf{\SuspectedLiberal (\sL)} \\
    Town Hall & CBS News \\
    & Chicago Tribune \\
    \textbf{\SuspectedConservative (\sC)} & CNN \\
    CS Monitor & Huffington Post \\
    Fox News & LA Times \\
    Washington Times & NY Times \\

        \bottomrule
    \end{tabular}
\caption{Labeled outlet categories.}
\label{tab:outlets}
\end{table}

\xhdr{Article length}
We expect the observed difference in reaction time between different categories
of outlets  to be reflected in the type of articles they publish.  The first
article feature we investigate is length in words,
shown in
Figure~\ref{fig:avg_article_length}. We
observe that declared outlets (especially liberal
ones) publish substantially longer articles; this difference is potentially
related to the longer time that these outlets take to cover the
respective speeches.

\xhdr{Fraction of quoted content}
To better understand the article-length differences, we examine the composition of
the articles in terms of quoted content.  In particular, we consider the
fraction of words in the article that are
quoted from a presidential speech.
Figure~\ref{fig:quoted_content} shows
that the (generally
shorter) \declaredconservative articles have a considerably higher proportion of
presidential content than most \declaredliberal articles, indicating a
different approach to
storytelling
that relies more on quotes and less on
exposition.

\xhdr{Summary}
Our exploration of the characteristics of a small set of labeled outlets reveals differences that go beyond the commonly studied \liberal-\conservative divide.  In particular, we find more substantial differences between self-declared outlets and suspected outlets in terms of their propensity of discussing Obama, their reaction time to new presidential speeches and the length of the articles citing these speeches.  Next we will investigate whether these differences also carry over to quoting patterns, and discuss how these observations further motivate an unsupervised approach to the study of media bias.

 %


\subsection{Outlet-to-quote graph analysis}

 We now explore the differences in the quoting
patterns of outlets from the four labeled categories.
To this end, we explore the structure of the bipartite graph $G$ connecting media outlets to
the quotes they cite (introduced in Section \ref{sec:graph}), focusing only the sub-graph induced by the labeled outlets.

We attempt to measure the
quoting pattern
similarity of
outlets from category $\catB$ to those from category $\catA$
as the likelihood of a source from category $\catB$
to cite a quote, given that the respective quote is also cited by
some outlet
in category $\catA$.
In terms of the outlet-to-quote graph $G$, we can quantify this as the
average proportion of quotes cited by outlets in $\catA$ that are also cited by
outlets in $\catB$:

$$\Prop_G(\catB | \catA) = \frac{1}{|o(\catA)|}\sum_{(u, v) \in o(\catA)} \frac{1}{|i(v)|} \sum_{(a, b) \in i(v)}\mathbf{1}[a \in \catB, u \neq a],$$

\noindent where $o(\catA)$ denotes the set of outbound edges in $G$ with the
outlet node residing in $\catA$, and $i(v)$ denotes the set of inbound edges in $G$
with $v$ as the destination quote node.  We will call $\Prop_G(\catB | \catA)$ the \textit{proportion-score} of $\catB$ given $\catA$.

\label{sec:surprise}
The proportion-score is not directly informative of the
quoting pattern
similarity
since it is skewed by differences in relative
sizes of the outlets in each category. 
To account for these effects,
we empirically estimate how unexpected $\Prop_G(\catB | \catA)$ is
given the observed degree distribution.
We construct random graphs by rewiring the edges of the original
bipartite graph~\cite{newman2001random}, such that for a large number of iterations we select edges
$u_1 \rightarrow v_1$ and $u_2 \rightarrow v_2$ to remove, where
$u_1 \neq u_2$ and $v_1 \neq v_2$, and replace these edges with
$u_2 \rightarrow v_1$ and $u_1 \rightarrow v_2$.

We use the randomly rewired graphs to build a hypothetical scenario where quoting happens at random, apart from trivial outlet-size effects.  We
can then quantify the deviation from this scenario using the  \textit{surprise} measure, which we defined as follows.
Let $R$ denote the set of all rewired graphs; given the original graph $G$ the
surprise $S_G(\catB | \catA)$ for categories $\catA$ and $\catB$ is:

$$S_G(\catB | \catA) = \frac{\Prop_G(\catB | \catA) - \mathbf{E}_{r \in
R}\Prop_r(\catB | \catA)}{\sqrt{\mathbf{Var}_{r \in R}\Prop_r(\catB | \catA)}}.$$

In other words, surprise measures the average difference between the proportion-score  calculated in the original graph $G$, and the one expected in the
randomly rewired graphs, normalized by the standard deviation. 
Surprise is, therefore, 
an asymmetric measure of similarity between the quoting patterns of outlets in
two given categories, that is not biased by the size of the outlets.

The surprise values between our four considered  categories are shown in Table \ref{tab:surprise}. A negative surprise score $S_G(\catB|\catA)$ indicates (in  units of
standard deviation) how much lower the proportion of quotes reported by outlets in $\catA$ that are also cited by outlets by $\catB$ is than in a hypothetical scenario where quotes are cited at random. 
For example, the fact that $S_G(\dC|\sC)$ is negative indicates that \declaredconservative outlets are much less likely to cite quotes reported by \suspectedconservative outlets than by chance, in spite of their suspected ideological similarity. Furthermore, we observe that \declaredliberal outlets are actually disproportionately more likely to cite quotes that are also reported by \declaredconservative outlets, in spite of their declared opposing ideologies.

\begin{table}[t]
    \centering
    \begin{tabular}{c  r r r r r}
    \toprule
        \multicolumn{1}{c}{} & \multicolumn{5}{ c }{\textbf{\catA}} \\
\sectionrule 
        \multicolumn{2}{c}{} & \multicolumn{1}{c}{\textit{\dC}} & \multicolumn{1}{c}{\textit{\sC}}  &\multicolumn{1}{c}{\textit{\sL}} & \multicolumn{1}{c}{\textit{\dL}} \\
        \multicolumn{1}{c}{} &\textit{\dC} & -3.5 & -6.1 & -9.7 & -3.4 \\
        \multicolumn{1}{c}{\textbf{\catB}} &\textit{\sC} & 0.7 & 1.4 & 0.1 & 1.0\\
        \multicolumn{1}{c}{} &\textit{\sL} & 1.1 & 5.6 & 6.1 & 3.5 \\
        \multicolumn{1}{c}{} &\textit{\dL} & 2.1 & -3.4 & 2.4 & -3.0 \\
        \bottomrule
    \end{tabular}
\caption {Surprise, $S_G(\catB | \catA)$, measures how much
more likely category \catB outlets are to cite quotes reported by \catA outlets,
compared to a hypothetical scenario where quoting is random.
}
\label{tab:surprise}
\end{table}

 Interestingly, for both categories of declared outlets, we find a high degree of within-category heterogeneity in terms of quoting patterns, with $S_G(\dC | \dC)$ and $S_G(\dL | \dL)$ being negative.  The reverse is true for suspected outlets: both $S_G(\sC | \sC)$ and $S_G(\sL | \sL)$ have positive values that indicate within category homogeneity (e.g., \suspectedliberal outlets are very likely to cite quotes that other \suspectedliberal outlets cite).  These observations bring additional evidence suggestive of the difference in nature between declared and suspected outlets.

\xhdr{Summary}  The surprise measure analysis not only confirms that there are systematic patterns in the underlying structure of the outlet-to-quote graph, but also shows that these patterns go beyond a naive \liberal-\conservative divide.  In fact, as also shown by our analysis of outlet and article characteristics, the declared-suspected distinction is often more salient.  These results emphasize the limitation of a naive supervised approach to classifying outlets according to ideologies: the outlets which we can confidently label as being \liberal or \conservative
are different in nature
from those that we would ideally like to classify.  This motivates our unsupervised framework for revealing the structure of political media coverage, which we discuss next.

\section{Large-scale Analysis}
\label{sec:discussion}


In this section we present a fully
unsupervised framework for
characterizing media bias. Importantly, this framework does not depend on predefined
dimensions of bias, 
and instead uncovers the 
structure of political media discourse directly from
quoting patterns.
In order to evaluate our model and confirm the systematicity of media coverage, %
we formulate the binary prediction task of whether a source will pick up a
quote.
We then use the low-rank embedding uncovered by 
our prediction method
to analyze and interpret
the emerging principal latent dimensions of bias and characterize them linguistically.

\subsection{Prediction: matrix completion}
\label{sec:matrixcompletion}
We attempt to model the latent dimensions that drive media coverage in a
predictive framework that we can objectively evaluate.  The task is to predict,
for a given media outlet and a given
quote from a presidential speech,
whether the outlet will choose
to report the quote or not.

Formally, 
we define $X = (x_{ij})$ to be the
outlet-by-quote adjacency matrix
such that
$x_{ij} = 1$ if\ outlet $i$ cites quote-cluster $j$ and $x_{ij} = 0$ otherwise.
In our task, we leave out a subset of the entries, and aim to recover them based on 
the other entries.

Inspired by recommender systems that reveal latent dimensions of user
preferences and item attributes, we use a {\em low-rank matrix completion}
approach.
By applying this methodology to news outlets and quotes, we
attempt to uncover
the dimensions along which quotes vary and along which news outlets manifest
their preference for certain types of quotes.

\begin{table}[t]
    \centering
    \begin{tabular}{l r r r r}
    \toprule
    \textbf{Method} & \textbf{P} & \textbf{R} & \textbf{F$_1$} & \textbf{MCC} \\
    \midrule
    quote popularity    & 0.07 & 0.29 & 0.11 & 0.12 \\ 
    + outlet propensity & 0.08 & 0.34 & 0.13 & 0.14 \\ 
    matrix completion   & {\bfseries 0.25} & 0.33 & {\bfseries 0.28} & {\bfseries 0.27} \\ 
    \bottomrule
    \end{tabular}
\caption{Classification performance of matrix completion,
    compared to the baselines
    in terms of precision,
    recall, $F_1$ score and Matthew's correlation coefficient.  Bold scores
are significantly better
(based on 99\% bootstrapped confidence intervals).}
\label{tab:matcomp}
\end{table}
\xhdr{Baselines}
We consider, independently, two baselines that do not take media bias into account: the popularity of a quote
$\mu^q_j = \mathbf{E}_{i} x_{ij}$ and the 
propensity of a news outlet to report
quotes from presidential speeches, $\mu^s_i = \mathbf{E}_j x_{ij}$ (where $\mathbf{E}$ is the sample mean).

A simple hypothesis is that quotes are cited only based on their newsworthiness, such that important quotes are cited more often:
$$\hat{x}_{ij} \propto \mu^q_j.$$

A step further is to also take into account the propensity of an outlet
to quote Obama at all:
$$\hat{x}_{ij} \propto \mu^q_j + \mu^s_i.$$

In a world without any media bias, this baseline 
would be very hard
to beat, since all outlets would cover the content proportionally
to its importance and to their own capacity, 
without showing
any
systematic
preference to any particular 
kind of content.

\xhdr{Low-rank approximation} 
More realistically, there are multiple dimensions that drive
media coverage.
To make use of this, we search for a low-rank
approximation $\hat{X} \approx \tilde{X}$, where $\tilde{X}$ is constructed
as follows:

We start by taking into account quote frequency in the weighted matrix $\bar{X} =
(\bar{x}_{ij})$ defined as:
$$\bar{x}_{ij} = \frac{x_{ij}}{\sqrt{\sum x_{:, j}}}.$$
Then, we build a row-normalized
$\tilde{X} = (\tilde{x}_{ij})$:
$$\tilde{x}_{ij} = \frac{\bar{x}_{ij}}{||\bar{X}_i||_2}.$$

We estimate $\hat{X}$ as to best reconstruct the observed values
(\emph{a}),
while keeping
the estimate low-rank by regularizing the $\ell_1$ norm of its singular values,
also known as its {\em nuclear norm}
(\emph{b})
\cite{mazumder2010spectral}:
$$ \operatorname{minimize}_{\hat{X}}
\underbrace{\frac{1}{2} || P_\Omega(\tilde{X}) - P_\Omega(\hat{X})||^2_F}_{(a)} +
\underbrace{\vphantom{\frac{1}{2}}\lambda ||\hat{X}||_*}_{(b)}, $$
where $P_\Omega$ is the
element-wise
projection over the space of
observed
values of $\tilde{X}$.  To solve the minimization problem, we use
a fast alternating least squares method \cite{hastie2014matrix}.

 \begin{figure}[t]
     \centering
     \includegraphics[width=0.48\textwidth]{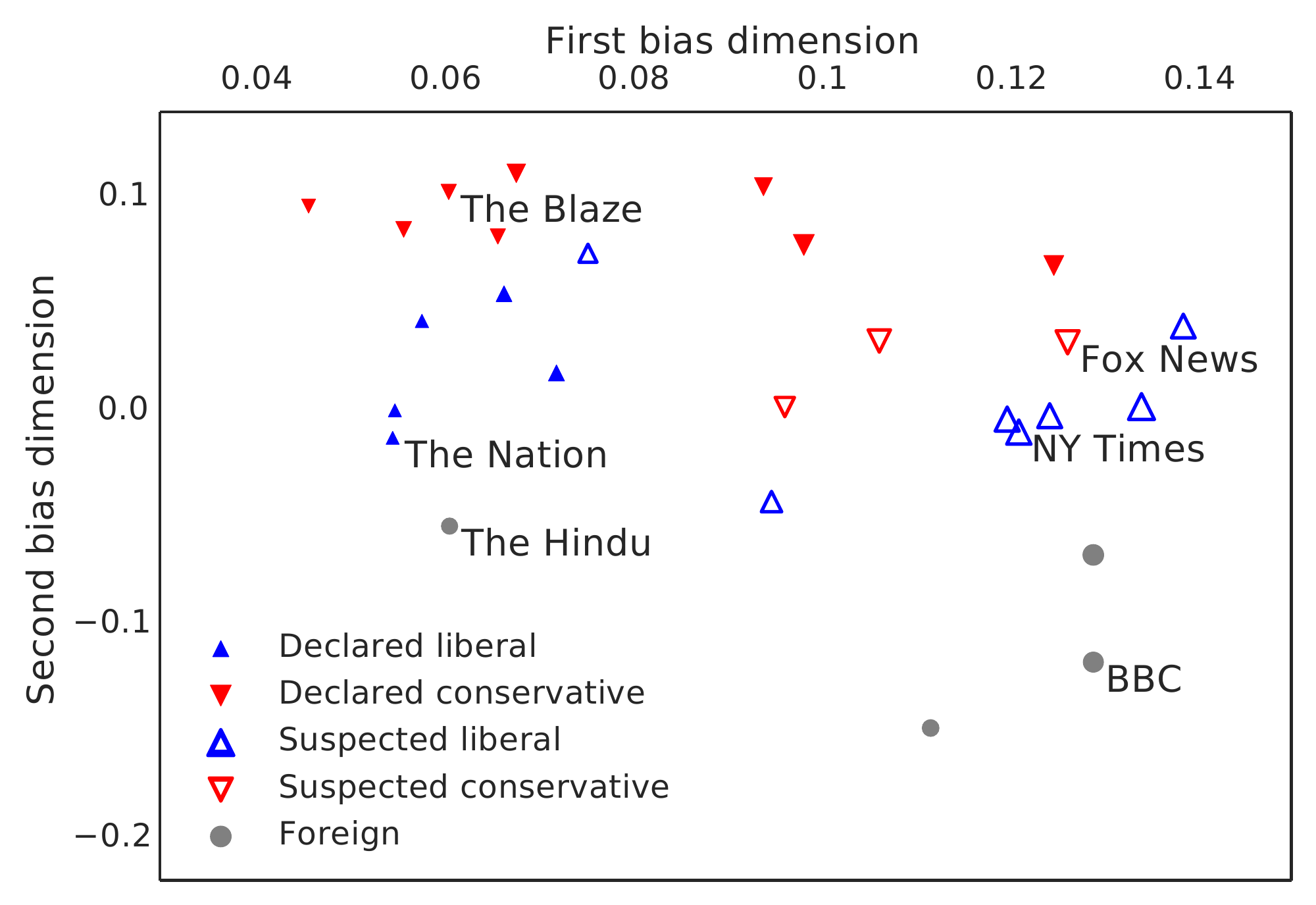}
     \caption{Projection of some of the media outlets
         onto the first two latent dimensions.  Filled and colored
markers         
 are outlets with self-declared political
 slant,
 such as {\em The Blaze} and {\em The Nation}, while unfilled
markers
are more popular outlets for which slants
 are
suspected, such as {\em Fox News} and the {\em
New York Times}.  Grey circles are international news outlets 
such as {\em BBC}
and {\em Hindu Times}.
Marker
sizes
are proportional to the propensity of
quoting Obama.}
\label{fig:firstcomps}
 \end{figure}
\begin{table}[t]
    \centering
    \small
    \begin{tabular}{l l l}
    \toprule
\textbf{High}    &   \textbf{Middle}  &   \textbf{Low} \\
\midrule
\href{http://weaselzippers.us/}{weaselzippers.us}   &   \href{http://www.motherjones.com/}{motherjones.com} &   \href{http://news.com.au/}{news.com.au} \\
\href{https://patriotpost.us/}{patriotpost.us}  &   \href{http://nypost.com/}{nypost.com}   &   \href{http://news.smh.com.au}{news.smh.com.au}  \\
\href{http://hotair.com/}{hotair.com}   &   \href{http://www.economist.com/}{economist.com} &   \href{http://ottawacitizen.com}{ottawacitizen.com}  \\
\href{http://www.freerepublic.com/}{freerepublic.com}   &   \href{http://www.macleans.ca/}{macleans.ca} &   \href{http://nationalpost.com}{nationalpost.com}    \\
\href{http://www.thegatewaypundit.com/}{thegatewaypundit.com}   &   \href{http://online.barrons.com/}{barrons.com}  &   \href{http://news.theage.com.au}{theage.com.au}    \\
\href{http://lonelyconservative.com/}{lonelyconservative.com}   &   \href{http://theplumline.whorunsgov.com}{whorunsgov.com}    &   \href{http://canada.com}{canada.com}    \\
\href{http://www.rightwingnews.com/}{rightwingnews.com} &   \href{http://www.csmonitor.com/}{csmonitor.com} &   \href{http://calgaryherald.com}{calgaryherald.com}  \\
\href{http://patriotupdate.com/}{patriotupdate.com} &   \href{http://www.cbsnews.com/}{cbsnews.com} &   \href{http://edmontonjournal.com}{edmontonjournal.com}  \\
\href{http://dailycaller.com/}{dailycaller.com} &   \href{http://www.latimes.com/}{latimes.com} &   \href{http://www.english.aljazeera.net/}{aljazeera.net} \\
\href{http://www.cnsnews.com/}{cnsnews.com} &   \href{http://www.cnn.com/}{cnn.com} &   \href{http://vancouversun.com/}{vancouversun.com}   \\
\href{http://www.wnd.com/}{wnd.com} &   \href{http://www.villagevoice.com/}{villagevoice.com}   &   \href{http://news.brisbanetimes.com.au}{brisbanetimes.com.au}  \\
\href{http://www.nationalreview.com/}{nationalreview.com}   &   \href{http://www.salon.com/}{salon.com} &   \href{http://montrealgazette.com}{montrealgazette.com}  \\
\href{http://www.americanthinker.com/}{americanthinker.com} &   \href{http://www.armytimes.com/}{armytimes.com} &   \href{http://bbc.co.uk}{bbc.co.uk}  \\
\href{http://www.theblaze.com/}{theblaze.com}   &   \href{http://democrats.org}{democrats.org}  &   \href{http://thesun.co.uk}{thesun.co.uk}    \\
\href{http://iowntheworld.com/}{iowntheworld.com} &   \href{http://tnr.com/}{tnr.com} &   \href{http://telegraph.co.uk}{telegraph.co.uk}  \\
\href{http://pjmedia.com/}{pjmedia.com} &   \href{http://rt.com}{rt.com}    &   \href{http://dailyrecord.co.uk}{dailyrecord.co.uk}  \\
\href{http://angrywhitedude.com/}{angrywhitedude.com}   &   \href{http://prnewswire.com}{prnewswire.com}    &   \href{http://independent.co.uk}{independent.co.uk}  \\
\href{http://ace.mu.nu/}{ace.mu.nu} &   \href{http://www.barackobama.com/}{barackobama.com} &   \href{http://theglobeandmail.com}{theglobeandmail.com}  \\

        \bottomrule
    \end{tabular}
\caption{Top-most, central and bottom-most news outlets according to the
second latent dimension.}
\label{tab:secondcomp}
\end{table}

\xhdr{Results}
We leave out 500,000 entries of the outlet-by-quote matrix
(out of 14.7 million) and divide them into equal
development and test sets.  The class distribution is heavily imbalanced, with
the positive class (quoting) occurring only about 1.6\% of the time.
In order to evaluate our model in a binary decision framework, we use Matthew's
correlation coefficient as the principal performance metric.  We tune the amount of
regularization $\lambda$ and the cutoff threshold on the development set.
The selected model has rank 3.
Table \ref{tab:matcomp} reports
the system's predictive
performance on the test set.
The latent low-rank model significantly outperforms both the quote popularity baseline
as well as the
baseline including outlet propensity, 
showing that the choices made by the media when covering political discourse
are not solely explained by newsworthiness and available space.
The performance of our model is twice that of the baselines in terms of
both $F_1$ and Matthew's correlation coefficient, and three times better in
terms of precision,
confirming that the latent quoting pattern bias is systematic and structured.
Motivated by our results, next we attempt to characterize the dimension of bias
with a spectral and linguistic analysis of the latent low-rank embedding.

\subsection{Low-rank analysis}\label{sec:lowrank}

\begin{table*}[t]
\centering
\textbf{First dimension of bias}
\begin{tabular}{p{1.5cm} p{15cm}}
\toprule
\textbf{High} & The principle that people of all faiths are welcome in this country, and will not be treated differently by their government, is essential to who we are. \\
\smallrule
& The United States is not, and will never be, at war with Islam. In fact, our partnership with the Muslim world is critical. \\
\smallrule
 & At a time when our discourse has become so sharply polarized [...] it's important for us to pause for a moment and make sure that we are talking with each other in a way that heals, not a way that wounds. \\
\sectionrule
\textbf{Low} & Tonight, we are turning the east room into a bona fide country music hall. \\
\smallrule
 & You guys get two presidents for one, which is a pretty good deal. \\
\smallrule
 & Now, nothing wrong with an art history degree---I love art history. \\
\bottomrule
\end{tabular}
\vspace{0.5cm}
\\
\textbf{Second dimension of bias}
\begin{tabular}{p{1.5cm} p{15cm}}
\toprule
\textbf{High} & Those of you who are watching certain news channels, on which
I'm not very popular, and you see folks waving tea bags around... \\
\smallrule
& If we don't work even harder than we did in 2008, then we're going to have a government that tells the American people, ``you're on your own.'' \\
\smallrule
& By the way, if you've got health insurance, you're not getting hit by a tax.\\
\sectionrule
\textbf{Middle} & Congress passed a temporary fix. A band-aid. But these cuts
are scheduled to keep falling across other parts of the government that provide
vital services for the American people. \\
\smallrule
& Keep in mind, nobody is asking them to raise income tax rates. All we're asking is for them to consider closing tax loopholes and deductions. \\
\smallrule
& The truth is, you could figure out on the back of an envelope how to get this done. The question is one of political will. \\
\sectionrule
\textbf{Low} & By the end of the next year, all U.S. troops will be out of Iraq.\\
\smallrule
& We come together here in Copenhagen because climate change poses a grave and growing danger to our people.\\
\smallrule
& Wow, we must come together to end this war successfully.\\
\bottomrule
\end{tabular}
\caption{Example quotes by President Obama mapped to top two dimensions of
quoting pattern bias.}
\label{tab:quotes}
\end{table*}

Armed with a low-rank space which captures the pre\-dict\-a\-ble quoting
behavior patterns of media,  we attempt to interpret the latent dimensions and gain
insights about these patterns.  This low-rank space is given by the singular value decomposition (SVD): $$\tilde{X}=USV^T,$$
\noindent where the rows of $U$ (respectively $V$) embed outlets (respectively quotes) in the latent space.

We start by looking at the 
mapping
 of the labeled outlets,
    as listed in Table \ref{tab:outlets},
in the space spanned by the latent dimensions.  Figure \ref{fig:firstcomps}
shows that 
the first two latent dimensions cluster the outlets in interpretable
ways. Outlets with
high values along the first axis appear to be more mainstream,
while outlets with
lower values more independent.\footnote{This
characterization is maintained after rewiring the bipartite graph in a way that
controls for the number of quotes that each outlet contributes to the data.
This dimension is therefore not
entirely
explained by outlet quoting propensity.}
Along the second dimension, declared \conservative outlets all have higher values
than declared \liberal outlets. International news outlets such as the BBC
and Al Jazeera have lower scores.
Going beyond our labeled outlets, we
look in detail at the projection of news outlets
along this dimension in Table~\ref{tab:secondcomp}, showing the outlets with the
highest and the lowest dimension
score,
as well as outlets in the middle.
A post-hoc investigation reveals that all the top outlets can be identified
with the \conservative ideology.  Furthermore, some of the highest ranked
outlets on this dimension are self-declared \conservative outlets that 
we did not consider in
 our manual categorization from Table~\ref{tab:outlets}, for example
\href{http://weaselzippers.us}{weaselzippers.us},
\href{http://hotair.com}{hotair.com} and
\href{http://rightwingnews.com}{rightwingnews.com}. The region around zero,
displayed in the second column of Table \ref{tab:secondcomp}, uncovers some
self-declared \liberal outlets not considered before, such as
\href{http://democrats.com}{democrats.com} and
President Obama's own blog, \href{http://barackobama.com}{barackobama.com},
as well as some outlets that are often accused of having a
\liberal slant, like \href{http://cbsnews.com}{cbsnews.com} and \href{http://latimes.com}{latimes.com}. Finally, the outlets
with the most negative scores around this dimension are all international
media outlets. 
 Given that our framework is completely unsupervised, we find the
alignment between the latent dimension and ideology to be surprisingly
strong.

\def\imagebox#1#2{\vtop to #1{\null\hbox{#2}\vfill}}

\begin{figure*}[t]
     \centering
 \subfigure[Sentiment]{
     \imagebox{55mm}{\includegraphics[width=0.28\textwidth]{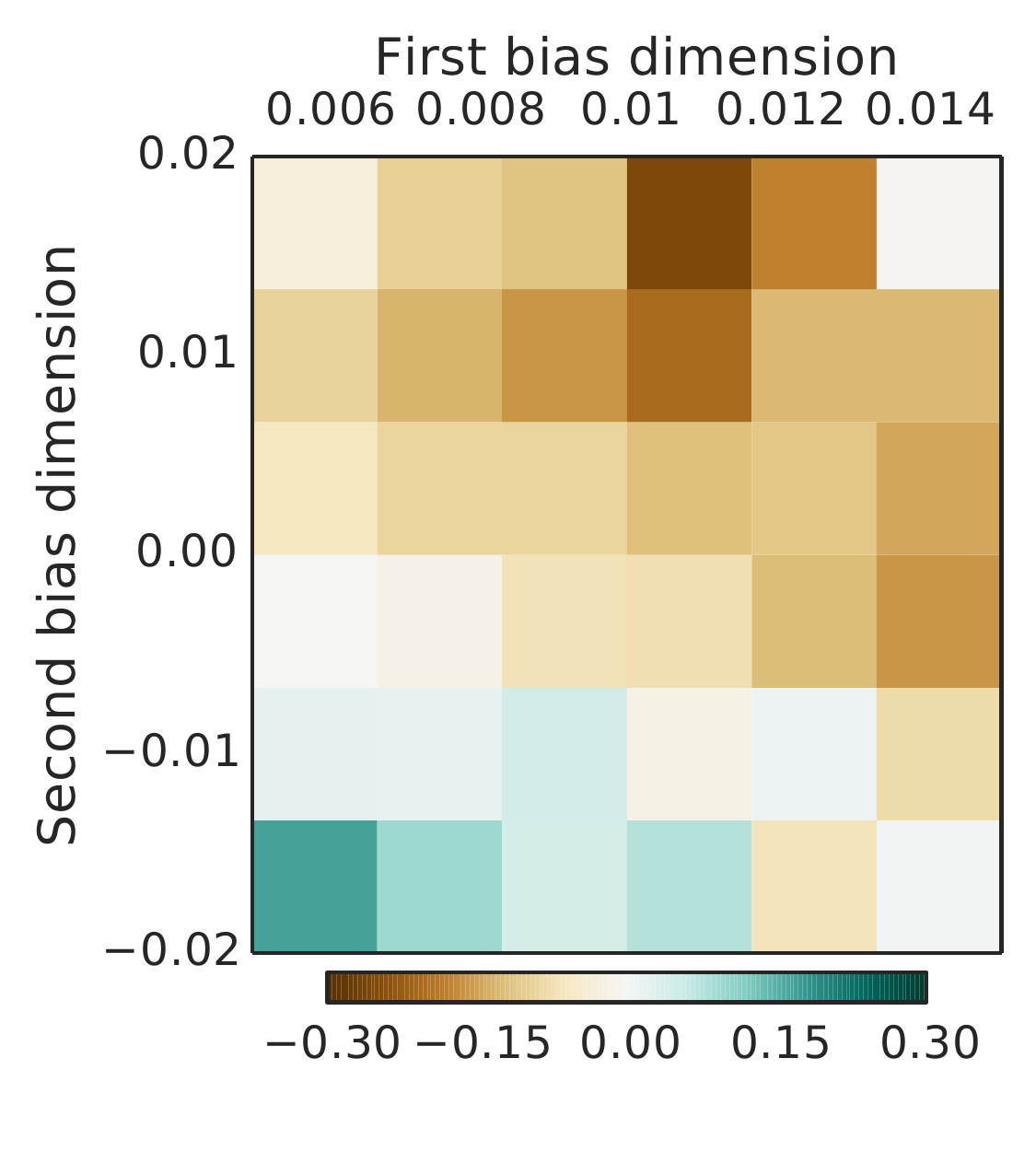}}
    \label{fig:sentiment}
}
\subfigure[Negation]{
    \imagebox{55mm}{\includegraphics[width=0.28\textwidth]{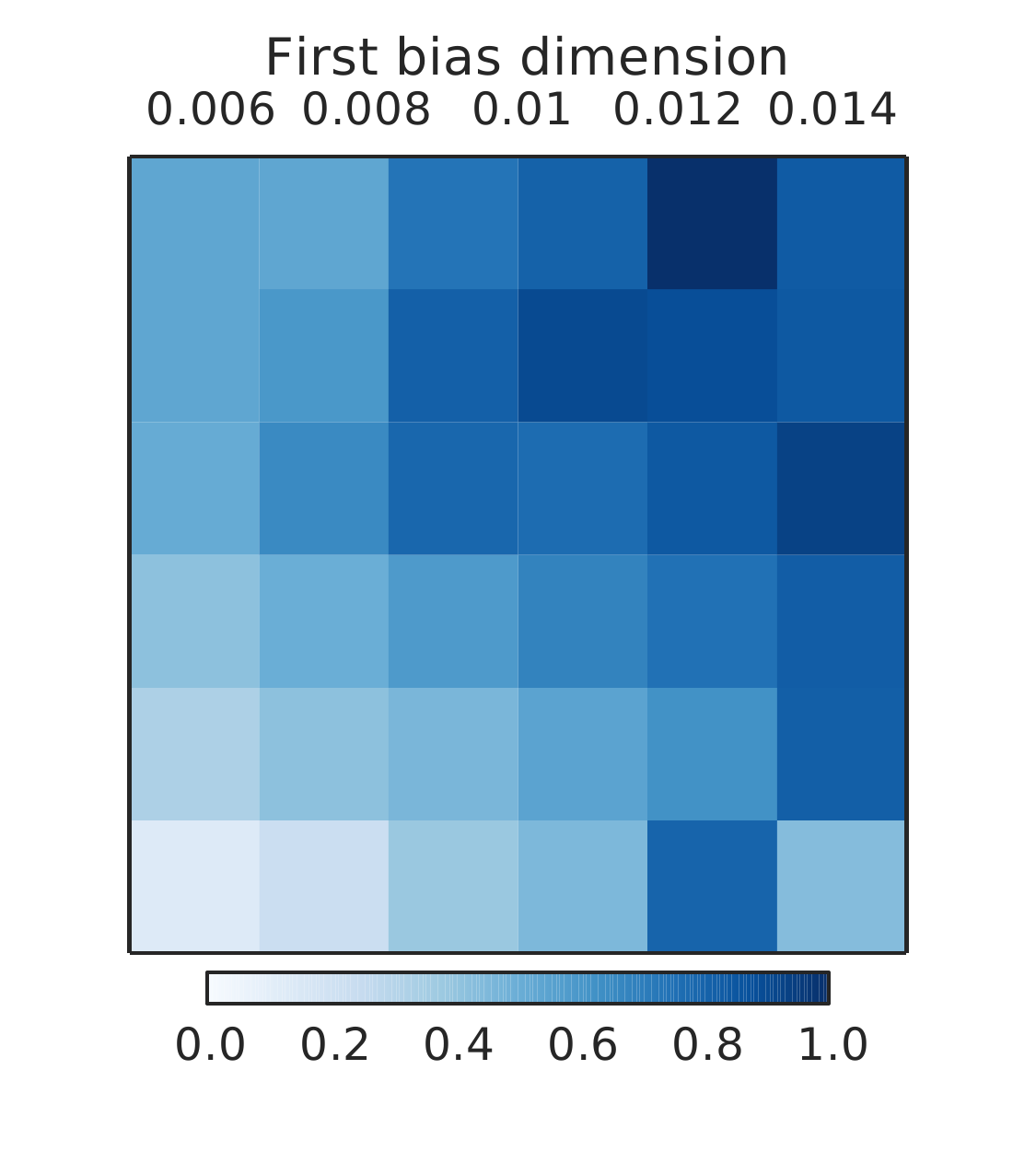}}
        \label{fig:negation}
}
\subfigure[Topics]{
    \imagebox{55mm}{\includegraphics[width=0.4\textwidth]{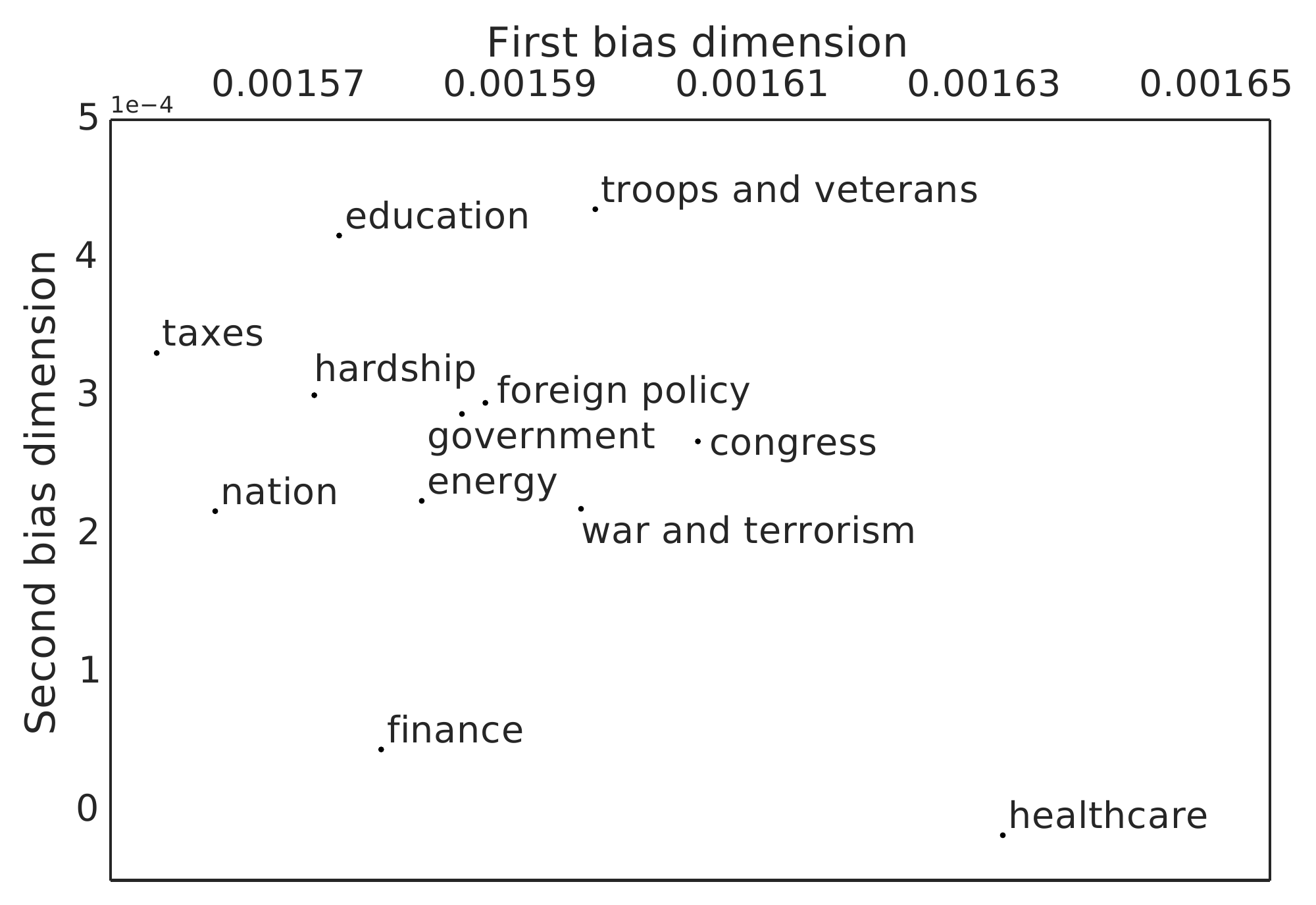}}
    \label{fig:topics}
}
\caption{Linguistic features projected on the first two latent dimensions: (a)
    sentiment of the quoted paragraphs; (b) proportion of quotes that contain
    negation; (c) dominant topics  discussed. 
Refer to
Figure~\ref{fig:firstcomps} for an interpretation of the latent space in terms of media outlet anchors.
}
 \end{figure*}

\subsection{Latent projection of linguistic features}
\label{sec:analysis}
So far we have seen that there are systematic differences in the quoting patterns
of different types of media outlets.
Since all the quotes we consider are from the same speaker, any systematic
linguistic differences that arise have the effect of portraying a different
persona of the president by different types of news outlets.
By choosing to selectively report 
certain kinds of quotes by Obama, outlets
are able to shape their audience's perception of how the president speaks and
what issues he chooses to speak about.

To characterize the effect of this selection, we make use of the fact that the
singular value decomposition of $\tilde{X}=USV^T$ provides not only a way
of mapping news outlets, but also presidential quotes to the
aforementioned latent space.
A selection of quotes that map to relevant
areas of the space (and that are cited at least five times) is shown in Table~\ref{tab:quotes}.

The quote embedding
allows us to perform a linguistic analysis of the presidential quotes and
interpret the results in the latent bias space.  Even though the latent representation is learned in
a completely language-agnostic way (starting
only from the outlet-quote graph), we find important language-related aspects.

\xhdr{Sentiment}
We applied Stanford's sentiment analyzer \cite{socher2013recursive} on the
presidential speeches and explore the relationship between the latent
dimensions and average sentiment of the paragraph surrounding the quote.
We find a negative correlation between the second dimension and sentiment
values: the quotes with high values along this dimension, roughly corresponding
to outlets ideologically aligned as \conservative, come from paragraphs with
more negative sentiment (Spearman $\rho=-0.32, p<10^{-7}$).  Figure
\ref{fig:sentiment} shows how positive and negative sentiment is distributed
along the first two latent dimensions. A diagonal effect is apparent,
suggesting that outlets
clustered in the international and independent region
portray a more positive Obama, while more mainstream
and
\conservative outlets
tend to reflect more negativity from the president's speeches.

\xhdr{Negation}
We also study how the presence of lexical negation (the word {\em not} and the
contraction {\em n't}) in a quote relates to the probability of media outlets from different regions of the latent bias space to cite that quote.  While lexical negation is in some cases related to sentiment,
it also corresponds to 
instances where the president contradicts or refutes a point, potentially relating to controversy.
 Figure \ref{fig:negation} shows the
likelihood of quotes to contain negation
in different areas of the latent space. 
The effect is similar:
quotes
with
negation seem more
likely in the 
region corresponding to mainstream \conservative outlets,
possibly because of highlighting the controversial aspects in the president's
discourse.

\xhdr{Topic analysis} We train a topic model using Latent Dirichet
Allocation \cite{blei2003latent} on all paragraphs from the
presidential speeches. We manually label the topics and discard the ones
related to the non-political ``background'' of the
speeches, such
as introducing other people and organizing question and answer sessions. 
We construct a topic--quote matrix $T = (t_{ij})$ such that $t_{ij} = 1$ if
the paragraph surrounding quote $j$
in the original speech
has topic $i$ as the
dominant
topic,\footnote{We define a topic to be dominant if its weight is larger by an
arbitrary threshold than the second highest weight.  We manually label the
topics using the ten most characteristic words.} and $0$ otherwise.  
We scale $T$ so that the rows (topics) sum to $1$, obtaining $\tilde{T}$, which we then project to the SVD latent space introduced earlier by solving for $L_T$ in $\tilde{T} = L_TSV^T$. Since V is orthonormal, the projection is given by $L_T = \tilde{T}VS^{-1}$.  Figure \ref{fig:topics}
shows the arrangement of the dominant topics in the latent space.
Quotes about the troops and war veterans are ranked on the top of the second
dimension, corresponding to more \conservative outlets, while financial and
healthcare quotes occupy the other end of the axis.  Healthcare is distanced
from other topics on the first axis, suggesting it 
is a topic of greater interest to mainstream news outlets rather than the
more
focused, independent media.

\xhdr{Lexical analysis} We attempt to capture a finer-grained linguistic
characterization of the space by looking at salient words and bigrams.  We
restrict our analysis to words and bigrams occurring in at least 100 and
at most 1000 quotes.  We construct a binary bag-of-words matrix $W$
where $(w_{ij}) = 1$ iff.\ word or bigram $i$ occurs in quote $j$.
Same
as with
topics, we scale the rows of $W$ (corresponding to word frequency) to obtain
$\tilde{W}$ and project onto the SVD latent space as $L_W = \tilde{W}VS^{-1}$.
Among the words that are projected highest on the first axis, we find
{\em republicans}, {\em cut}, {\em deficit} and {\em spending}. Among the center
of the ranking we find words and phrases
such as {\em financial crisis}, {\em
foreign oil}, {\em solar}, {\em small business}, and {\em Bin Laden}. The
phrase {\em chemical weapons} also appears near the middle, possibly as
an effect of \liberal outlets being critical of the decisions former Bush administration.  On the negative
end of the spectrum, corresponding to international outlets, we find words
such as {\em countries}, {\em international}, {\em relationship}, {\em alliance}
and country names such as {\em Iran}, {\em China}, {\em Pakistan}, and {\em Afghanistan}.

Overall, the mapping of linguistic properties of the quotes in the latent bias space is surprisingly consistent, and suggest that outlets in different regions of this space consistently portray different presidential personae to their audience.

\section{Further Related Work}
\label{sec:related}


\xhdr{Media bias} Our work relates to an extensive body of literature---spanning across political science, economics and communication studies---that gives
theoretical and empirical accounts of media bias and its effects.  We refer the reader to a recent comprehensive survey 
of media bias 
\cite{Prat::2013}, and focus here on the studies that are most relevant to our approach.

\xhdr{Selection patterns} Several small-scale studies investigate 
subjects that 
media outlets select to cover by relying on hand annotated slant labels.   For instance, by tracing the media coverage of 32 hand-picked scandal stories, it was shown that outlets with a 
\conservative slant are more likely to cover scandals involving \liberal politicians, and vice-versa  \cite{puglisi2011newspaper}.  Another study \cite{baum2008new} focuses on 
the choices that five online news outlets make with respect to which stories to display in their top news section,
 and reports that \conservative outlets show signs of partisan filtering.  In contrast, by relying on an unsupervised methodology, our work explores selection patterns in data involving orders of magnitude more selector agents and items to be selected.
Closer to our approach are methods that show political polarization starting from linking patterns in blogs  \cite{adamic2005political} or from the structure of the retweet graph in Twitter \cite{Conover:IeeeXplore:2011}.  These approaches operate on a predefined \liberal-\conservative dimension, and assume available political labels.  Furthermore, the structure they exploit does not directly apply to the setting of news media articles.

\xhdr{Language and ideology} 
Recently, natural language processing techniques 
 were applied to identify ideologies in a variety of large scale text collections, including congressional debates \cite{iyyerpolitical,nguyen2013lexical}, presidential debates \cite{Lin:MachineLearningAndKnowledgeDiscoveryInDatabases:2008}, academic papers \cite{Jelveh:Emnlp:2014}, books \cite{sim2013measuring},  and Twitter posts
\cite{cohen2013classifying,Volkova:ProceedingsOfTheAcl:2014,wong2013quantifying}.
 All this work operates on a predefined dimension of
\conservative--\liberal
political ideology using known slant labels; in the news media domain
slant is seldom declared or proven with certainty and
thus we need to resort to an unsupervised methodology.

\xhdr{Quote tracking} 
Recent work has focused quoting practices \cite{Soni:ProceedingsOfAcl:2014} and on the task of efficiently tracking and matching quote snippets as they evolve, both over a set period of time \cite{memetracker2009}, as well as over an longer, variable period of time \cite{nifty2012}. We adapt this task in order to news article quotes with presidential speech segments and build our outlet-to-quote graph.

\vspace{0.1in}
\section{Conclusion}
\label{sec:conclusion}


We propose an unsupervised framework for uncovering and characterizing media bias starting from quoting patterns.  We apply this framework to a dataset of matched news articles and presidential speech transcripts, which we make publicly available together with an online visualization that can facilitate further exploration.

There is systematic bias in the quoting patterns of different types of news
sources.  We find that the bias goes beyond simple newsworthiness and space limitation effects, and we objectively
quantify
this by showing our model to be predictive of quoting activity,
without making any {\em a priori} assumptions regarding the dimension of bias and without requiring labeling of the news domains.
When 
comparing the unsupervised model with self-declared political slants, we
find that an important dimension of bias is roughly aligned with an ideology
spectrum ranging from \conservative, passing through \liberal, to the international
media outlets.

By selectively choosing to report certain types of quotes by the same speaker,
the media has the power to portray different personae of the speaker.
Thus, 
an
audience only following one type of
media may witness a presidential persona that is different 
from the
one portrayed by other types of media or from what the president tries to project.  By conducting a linguistic analysis on the latent dimensions revealed by our framework,
we find that differences go beyond topic selection, and that \mainstream \conservative outlets portray a persona that is characterized by negativism, both in terms of negative sentiment and in use of lexical negation.

\xhdr{Future work}
Throughout our analysis, we don't take into account potential changes
in a media outlet's behavior over time. Modeling temporal effects could reveal idealogical
shifts and differences in issue framing.

Furthermore, natural language techniques can be better tuned to insights from political science in order to produce  tools and resources more suited for analyzing 
political discourse \cite{Grimmer:PoliticalAnalysis:2013}.  
For instance, exploratory analysis shows that state-of-the-art
sentiment analysis tools fail to capture 
subtle nuances
in political commentary and we expect that fine-grained opinion mining
can achieve this better
\cite{Li:Emnlp:2014a}. 

Finally, news sources often take the liberty 
of skipping or altering certain words when
quoting.  While these changes are often made to improve
readability,
we
speculate
that systematic patterns in such edits could uncover
different dimensions of media bias.

\vspace{0.25in}

\section*{Acknowledgements}
We would like to acknowledge the valuable input provided by
Mathieu Blondel, %
Flavio Chierichetti,
Justin Grimmer,
Scott Kilpatrick,
Lillian Lee,
Alex Niculescu-Mizil,
Fabian Pedregosa, %
Noah Smith and
Rebecca Weiss.  We are also thankful to the anonymous reviewers for their valuable comments and suggestions.  This research has been supported in part by 
a Google Faculty Research Award,
NSF IIS-1016909,              
NSF IIS-1149837,       
NSF IIS-1159679,              
NSF CNS-1010921,              
DARPA SMISC,
 Boeing,                    
 Facebook,
 Spinn3r,
Volkswagen,                 
and Yahoo.

\vfill\eject

\end{document}